\documentclass[10pt, a4paper]{article}

\usepackage[final]{lrec-coling2024} 
\usepackage{multirow}
\usepackage{arydshln}
\usepackage{xcolor}

\title{Automatic Construction of a Large-Scale Corpus for Geoparsing Using Wikipedia Hyperlinks}

\name{
    Keyaki Ohno$^{1}$ \quad
    Hirotaka Kameko$^{2}$ \quad 
    Keisuke Shirai$^{1}$ \\
    {\bf \large Taichi Nishimura\thanks{$^*$Work done while the author was at Kyoto University.}$^{*3}$} \quad 
    {\bf \large Shinsuke Mori$^{2}$}} 

\address{
    $^{1}$Graduate School of Informatics, Kyoto University, Japan \\
    $^{2}$Academic Center for Computing and Media Studies, Kyoto University, Japan \\
    $^{3}$LY Corporation, Japan \\
    \{ohno.keyaki.57r, shirai.keisuke.l82\}@kyoto-u.jp \\
    \{kameko,forest\}@i.kyoto-u.ac.jp \quad tainishi@lycorp.co.jp
}

\abstract{
Geoparsing is the task of estimating the latitude and longitude (coordinates) of location expressions in texts. Geoparsing must deal with the ambiguity of the expressions that indicate multiple locations with the same notation. For evaluating geoparsing systems, several corpora have been proposed in previous work. However, these corpora are small-scale and suffer from the coverage of location expressions on general domains. In this paper, we propose \textit{Wikipedia Hyperlink-based Location Linking} (WHLL), a novel method to construct a large-scale corpus for geoparsing from Wikipedia articles. WHLL leverages hyperlinks in Wikipedia to annotate multiple location expressions with coordinates. With this method, we constructed the WHLL corpus, a new large-scale corpus for geoparsing. The WHLL corpus consists of 1.3M articles, each containing about 7.8 unique location expressions. 45.6\% of location expressions are ambiguous and refer to more than one location with the same notation. In each article, location expressions of the article title and those hyperlinks to other articles are assigned with coordinates. By utilizing hyperlinks, we can accurately assign location expressions with coordinates even with ambiguous location expressions in the texts. Experimental results show that there remains room for improvement by disambiguating location expressions.
 \\ \newline }

\newcommand{\tabref}[1]{{\tableautorefname~\ref{#1}}}
\newcommand{\figref}[1]{{\figureautorefname~\ref{#1}}}

\newcommand{\colora}{FF2800}
\newcommand{\colorb}{005AFF}

\newcommand{\xtc}[2]{\textcolor[HTML]{#1}{#2}}

\newcommand{\colory}{106802}
\newcommand{\colorz}{fc5c00}

\begin{document}
\maketitleabstract

\section{Introduction}
Recognizing the spatial information indicated by location expressions in texts is a promising direction in text comprehension by machines. This enables us to automatically analyze large amounts of texts worldwide and read the texts about places of interest. In addition, it would be possible to organize texts based on their geographic characteristics.

Geoparsing is the task of estimating the latitude and longitude (coordinates) of location expressions in texts. Typically, geoparsing consists of two tasks: recognizing location expressions in texts (geotagging) and estimating their coordinates (geocoding). One of the challenges in geoparsing is the disambiguation of the expressions. For example, the word \textit{Melbourne} is ambiguous as it could be a city in Australia, the United Kingdom, Canada, or the United States. We focus on such ambiguous location expressions where the same notation indicates multiple locations.

Previous work proposed corpora to evaluate the performance of disambiguation by geoparsing systems~\citep{LGL, WikToR, GeoVirus}. These corpora are relatively small-scale (e.g., hundreds or thousands) or focus on specific domains. Thus, using these corpora suffers from the coverage of location expressions when evaluating the systems on general domains. Therefore, a large-scale corpus containing general articles is required to alleviate these issues. 

\begin{figure}[t]
\begin{center}
\includegraphics{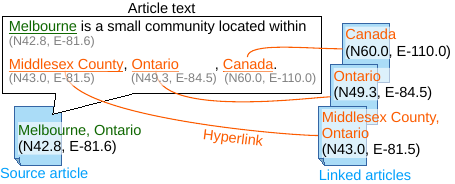}
\caption{Coordinates annotation using hyperlinks. 
    \xtc{\colorz}{Location expressions with hyperlinks} are annotated with the coordinates of the linked articles, and \xtc{\colory}{those equal to the article title} are annotated with the coordinates of the article.
    A sample article of our corpus is shown in \figureautorefname~\ref{fig-sample}.
    } 
\label{fig-intro}
\end{center}
\end{figure}

In this paper, we propose \textit{Wikipedia Hyperlink-based Location Linking} (WHLL) that automatically assigns multiple location expressions with coordinates in Wikipedia articles. Contrary to previous work~\cite{WikToR}, WHLL realizes the coordinates annotation to location expressions other than the article title by leveraging hyperlinks in Wikipedia dump files. Note that WHLL does not predict any coordinates and uses those registered in Wikipedia articles. \figref{fig-intro} shows a schematic of WHLL. Based on this method, we constructed the WHLL corpus, a new large-scale corpus for geoparsing. It has more than 1.3M articles with about 7.8 unique location expressions per article, and 45.6\% of total location expressions are ambiguous. Our code and data are publicly available.\footnote{\url{http://www.lsta.media.kyoto-u.ac.jp/resource/data/WHLL/home-e.html}}

\section{Related Work}\label{sec-related}
\subsection{Methods for geoparsing}
Various methods have been proposed in terms of disambiguation. This includes rule-based methods that refer to population, geographic distance between candidates or administrative level on a database~\citep{aldana-bobadilla2020adaptive}, statistical methods that calculate probability distributions from data in advance~\citep{speriosu2013text, delozier2015gazetteer-independent}, and machine learning-based methods~\citep{Kurohashi-SVM, GeoVirus, Multi-Level}.

\subsection{Corpora for geoparsing}\label{RW}
The evaluation of geopasing, especially the task of estimating coordinates for each location expression in a text, has often relied on corpora such as Local-Global Lexicon (LGL)~\citep{LGL}, Wikipedia Toponym Retrieval (WikToR)~\citep{WikToR} and GeoVirus~\citep{GeoVirus}. \textbf{LGL} consists of 588 news articles. Each location expression was manually annotated. They collected newspapers located near locations whose names were ambiguous, such as the Paris News (Texas) and the Paris Post-Intelligencer (Tennessee). \textbf{WikToR} is a corpus created from 5,000 Wikipedia pages. They selected ambiguous location expressions using GeoNames and collected the Wikipedia first paragraphs of these locations using GeoNames API. Since each article in the corpus is annotated only with the coordinates of the article title, it has only one unique coordinate. \textbf{GeoVirus} consists of 229 WikiNews about global disease outbreaks and epidemics. Coordinates were manually determined, referring to Wikipedia or GeoNames.

In addition to evaluation data, machine learning-based geoparsing requires a considerable amount of training data. Some previous studies actually used more than 1 million Wikipedia pages for machine learning. 

It is desirable that a corpus for geoparsing (i) contains many ambiguous location expressions and (ii) is unbiased in its sampling of locations. Furthermore, especially as training data, it is desirable that (iii) multiple location expressions appear within a text. This is because it is useful for learning the geographical relationships between each location expression. However, a large-scale corpus that meets these requirements has not been published to date. Corpora such as those mentioned above are biased in their sampling, such as enriching ambiguous location expressions, for efficient evaluation. Therefore, we created a corpus that satisfies these three requirements. \tableautorefname~\ref{table-comparison} shows that our corpus is large and contains ambiguous location expressions comparable to those of previous corpora.

\begin{table}[t]
\centering
\begin{tabular}{lrc}

\hline
       &                                & \#unique location\\
Corpus & \multicolumn{1}{c}{\#articles} & expressions\\
       &                                & par article\\
\hline
LGL         &        588 & 4.2 \\
WikToR      &      5,000 & 1.0 \\
GeoVirus    &        229 & 6.6 \\\cline{1-3}
WHLL corpus &  1,315,117 & 7.8 \\
\hline

\end{tabular}
\caption{Comparison of the WHLL corpus and previously proposed corpora for geoparsing.}
\label{table-comparison}
\end{table}

\subsection{Automatic dataset creation from Wikipedia metadata}
Outside the field of geoparsing, there are various works on automatic dataset creation using Wikipedia metadata. \citet{yang2014semi} utilized entity tags in Wikipedia as natural annotations for word segmentation tasks. \citet{ling2021fine} leveraged Wikipedia text and its hyperlinks to create training data for NER. \citet{pan2017cross} proposed to link non-English name mentions to an English knowledge base automatically based on Wikipedia anchor links.

\section{Wikipedia Hyperlink-based Location Linking}\label{sec-whll}
\begin{figure*}[t]
\centering
\small
\begin{tabular}{llll}
\hline
Title & \multicolumn{3}{l}{Melbourne, Ontario}\\\hline

\multirow{4}{*}{Text} & \multicolumn{3}{l}{\multirow{4}{133mm}{\xtc{\colorb}{\underline{Melbourne}} is a small community located within \xtc{\colorb}{\underline{Middlesex County}}, \xtc{\colorb}{\underline{Ontario}}, \xtc{\colorb}{\underline{Canada}}. It lies on the boundary between two municipalities, \xtc{\colorb}{\underline{Strathroy-Caradoc}} and \xtc{\colorb}{\underline{Southwest Middlesex}}. About half the population of \xtc{\colorb}{\underline{Melbourne}} lives in each municipality. The community was probably named for \xtc{\colora}{\underline{Melbourne}}, \xtc{\colora}{\underline{Victoria}}, Australia.}}\\

& \\
& \\
& \\\hline
\multirow{10}{*}{Annotation} & Span \hspace{15mm} & Notation in the text \hspace{15mm} & (Latitude, Longitude) \\\cdashline{2-4}

    & (0, 9)     & ``\xtc{\colorb}{Melbourne}''           & (\phantom{-}42.81667, \phantom{0}-81.55194) \\
    & (46, 62)   & ``\xtc{\colorb}{Middlesex County}''    & (\phantom{-}43.00000, \phantom{0}-81.50000) \\
    & (64, 71)   & ``\xtc{\colorb}{Ontario}''             & (\phantom{-}49.25000, \phantom{0}-84.50000) \\
    & (73, 79)   & ``\xtc{\colorb}{Canada}''              & (\phantom{-}60.00000, -110.00000) \\
    & (133, 150) & ``\xtc{\colorb}{Strathroy-Caradoc}'' & (\phantom{-}42.95750, \phantom{0}-81.61667) \\    
    & (155, 174) & ``\xtc{\colorb}{Southwest Middlesex}'' & (\phantom{-}42.75000, \phantom{0}-81.70000) \\
    & (205, 214) & ``\xtc{\colorb}{Melbourne}''           & (\phantom{-}42.81667, \phantom{0}-81.55194) \\
   & (280, 289) & ``\xtc{\colora}{Melbourne}''           & (-37.81417, \phantom{-}144.96306) \\
   & (291, 299) & ``\xtc{\colora}{Victoria}''           & (-37.00000, \phantom{-}144.00000) \\
    \hline
    
\end{tabular}
\caption{A sample article from our corpus (including 7 \xtc{\colorb}{Canadian} and 2 \xtc{\colora}{Australian} location expressions). Annotation includes string span, notation, latitude, and longitude. The last \textit{\xtc{\colora}{Melbourne}} in the HTML text of this article had a hyperlink to another Wikipedia article, resulting in different coordinates annotated. Note that location expressions without hyperlinks (e.g., \textit{Australia} in the sample) are ignored in the annotation.}
\label{fig-sample}
\end{figure*}

Wikipedia Hyperlink-based Location Linking (WHLL) is a novel method to automatically assign multiple location expressions in Wikipedia articles with coordinates. Here, we focus on articles of locations. Based on WHLL, we create a new corpus, the WHLL corpus. In this section, we first describe WHLL in Section~\ref{sub:AA} and then introduce the WHLL corpus and its statistics in Section~\ref{sub:corpus}.

\subsection{Construction process}
\label{sub:AA}
Many articles on Wikipedia are related to locations, and each article contains the single coordinates of the location. The articles contain more location expressions than the title, and several expressions are annotated with hyperlinks to other articles. Thus, location expressions in the article can be divided into two types: expressions with and without hyperlinks. WHLL automatically assigns these location expressions with coordinates of the original article and those of the linked articles. 

The process consists of two steps: identifying the location expressions' span in the article and assigning coordinates to the identified expressions. For the expressions with hyperlinks, WHLL identifies the span of the expressions based on special tags (e.g., in HTML dump files, such expressions are surrounded by \textsf{<a>} tag). The expressions are then assigned the coordinates of the articles to which the expression hyperlinks.

For the expressions without hyperlinks, WHLL targets expressions of the article title and identifies their spans. More concretely, WHLL searches for expressions matching any of the following:
\begin{itemize}
    \item The article title (e.g., \textit{``Melbourne, Ontario''}),
    \item String obtained by removing the comma and the right side from the title (e.g., \textit{``Melbourne''} from \textit{``Melbourne, Ontario''}),
    \item String obtained by removing the parenthesis and its content from the title (e.g., \textit{``Waterloo''} from \textit{``Waterloo (Albertson, North Carolina)''}).
\end{itemize}
The matched expressions are then assigned with the coordinates annotated on the current article.

WHLL utilizes two types of Wikipedia dump files: an HTML dump for obtaining formatted articles and a CirrusSearch dump for realizing the assignments of the expressions with coordinates based on hyperlinks. Since WHLL is automatic, it provides consistent annotations and can be applied to large texts with minimum cost. Further, one can update or newly create the corpus based on dump files. Note that WHLL constructs the corpus from Wikipedia articles; it does not newly predict coordinates. 

\subsection{The WHLL corpus}
\label{sub:corpus}
The WHLL corpus is based on two dump files: the Wikipedia HTML dump file released on July 1, 2023\footnote{\url{https://dumps.wikimedia.org/other/enterprise_html/runs/20230701/enwiki-NS0-20230701-ENTERPRISE-HTML.json.tar.gz}. Accessed on July 17, 2023 and checked the expiration on October 20, 2023.} and the Wikipedia CirrusSearch dump file released on July 10, 2023.\footnote{\url{https://dumps.wikimedia.org/other/cirrussearch/20230710/enwiki-20230710-cirrussearch-content.json.gz}. Accessed on July 11, 2023 and checked the expiration on October 20, 2023.} We use only articles that are annotated with coordinates. \figureautorefname~\ref{fig-sample} shows a sample article in the corpus.

\begin{figure}[t]
\begin{center}
\includegraphics[width=77mm]{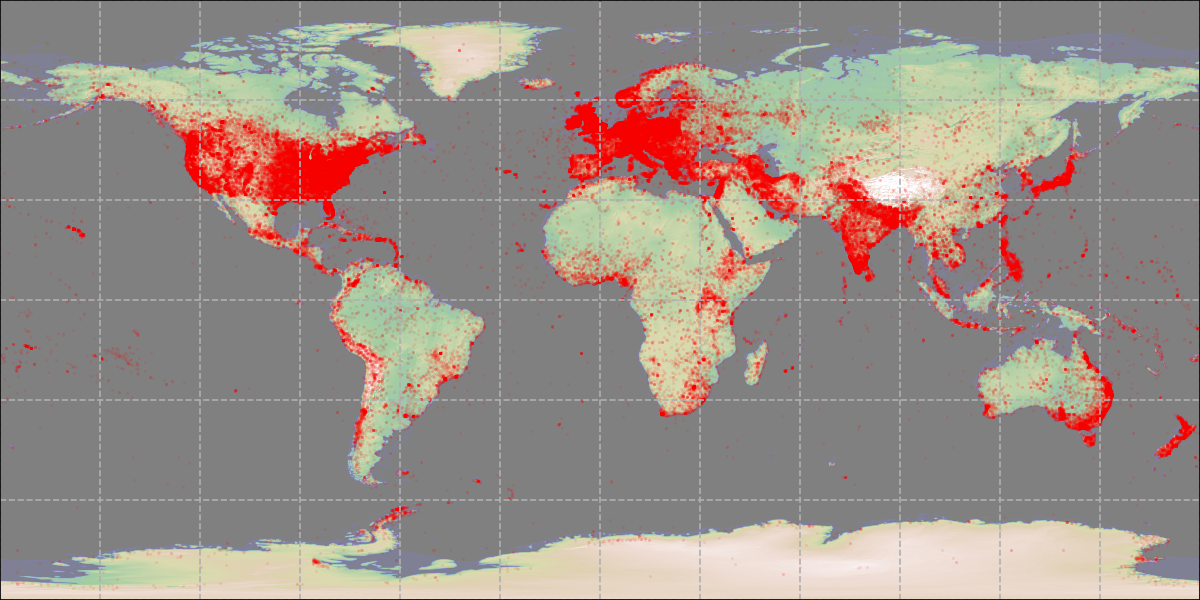}
\caption{Coordinates that appear more than 10 times in the WHLL corpus.}
\label{fig-map}
\end{center}
\end{figure}

\tabref{table-statistics} shows the statistics of the WHLL corpus. The WHLL corpus has 1.3M articles and over 14.7M location expressions (1.6M unique expressions). 45.6\% of the expressions are ambiguous: there are multiple coordinates corresponding to the single notation. In addition, 9.9\% of the expressions are recessive: they are ambiguous expressions and do not refer to the most frequent coordinates. Disambiguating these expressions is challenging because simply choosing the most frequent coordinates fails to find the correct coordinates. Each article contains 7.8 unique location expressions on average, indicating more dense annotations than WikToR~\cite{WikToR}. As illustrated in \figref{fig-map}, the expressions in the WHLL corpus refer to locations around the world. 

\begin{table}[t]
\centering
\begin{tabular}{lr}
\hline \hline
\multicolumn{2}{c}{\textbf{Total}} \\
\hline
\#articles & 1,315,117 \\
\#sentences & 23,334,035 \\
\#location expressions & 14,726,908 \\
\quad (Ambiguous & 45.6\%) \\
\quad (Ambiguous \& Recessive & 9.9\%) \\
\#unique location expressions & 1,571,291 \\
\quad (Ambiguous & 8.1\%) \\ 

\hline \hline

\multicolumn{2}{c}{\textbf{Per article}} \\
\hline
\#sentences& 17.7 \\
\#tokens & 420.1 \\
\#location expressions & 11.3 \\
\#unique location expressions & 7.8 \\
\hline \hline
\end{tabular}
\caption{Statistics of our corpus. Ambiguous means the location expressions indicate multiple coordinates with the same notation.}
\label{table-statistics}
\end{table}

\section{Experiments}\label{sec-experiments}
We conducted geocoding experiments to show the necessity of the disambiguation for the WHLL corpus. Here, we used ground-truth location expressions by assuming that geotagging was already finished.\footnote{In previous work~\citep{LGL,GeoVirus}, geotagging was solved by using named entity recognition models.} In this section, we first describe our method for geocoding and experimental settings and then report our results.

\subsection{Method}
We used GeoNames\footnote{\url{https://www.geonames.org/}. Accessed on October 20, 2023.} as a geographic database. Each entry in GeoNames has a name, a list of alternate names, coordinates and other information. We assigned an entry to each location expression by following strategies.

\textbf{Familiarity-based strategy} is the one that chooses an entry whose name matches the focused location expression and having the highest number of alternate names registered. This is based on our observation that famous locations tend to have many alternate names in GeoNames. If no candidates are found in the database, the coordinates just before in the sentence are copied.

\begin{figure}[t]
\begin{center}
\includegraphics[width=77mm]{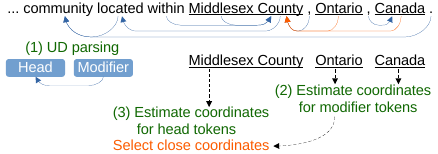}
\caption{Dependency-based strategy.}
\label{fig-UD}
\end{center}
\end{figure}

\textbf{Dependency-based strategy} is the one that chooses an entry considering the dependencies between location expressions. We assume that the locations indicated by two dependent location expressions are geographically close. Using Universal Dependencies (UD)~\citep{mcdonald2013universal} as the dependency structure, we selected coordinates by the following procedure (\figureautorefname~\ref{fig-UD}):
\begin{enumerate}
\renewcommand{\labelenumi}{(\arabic{enumi})}
    \item Find mutually related location expressions using dependency structure between tokens.
    \item For the modifier tokens, choose candidates (coordinates) in the same way as familiarity-based strategy. These location expressions tend to provide supplementary information in the sentence and indicate relatively large geographic locations.
    \item For the head tokens, select the closest coordinates to (2) from the candidates or copying (2) if no candidates are found. These location expressions tend to provide the main information of the sentence and indicate relatively small geographical locations, often included in the location of (2).
\end{enumerate}

\subsection{Experimental settings}
We used 10,000 randomly selected articles from the created corpus for evaluation. We used Stanford NLP Group's stanza~\citep{stanza} as the tokenizer and the dependency parser~\citep{dozat2017deep}. Following previous work~\cite{GeoVirus,Multi-Level}, we used accuracy@161 km as an evaluation metric. This is the proportion of location expressions whose coordinates were estimated within 161 km of their actual coordinates.

\subsection{Results} 
\begin{figure}[t]
\begin{center}
\includegraphics[width=77mm]{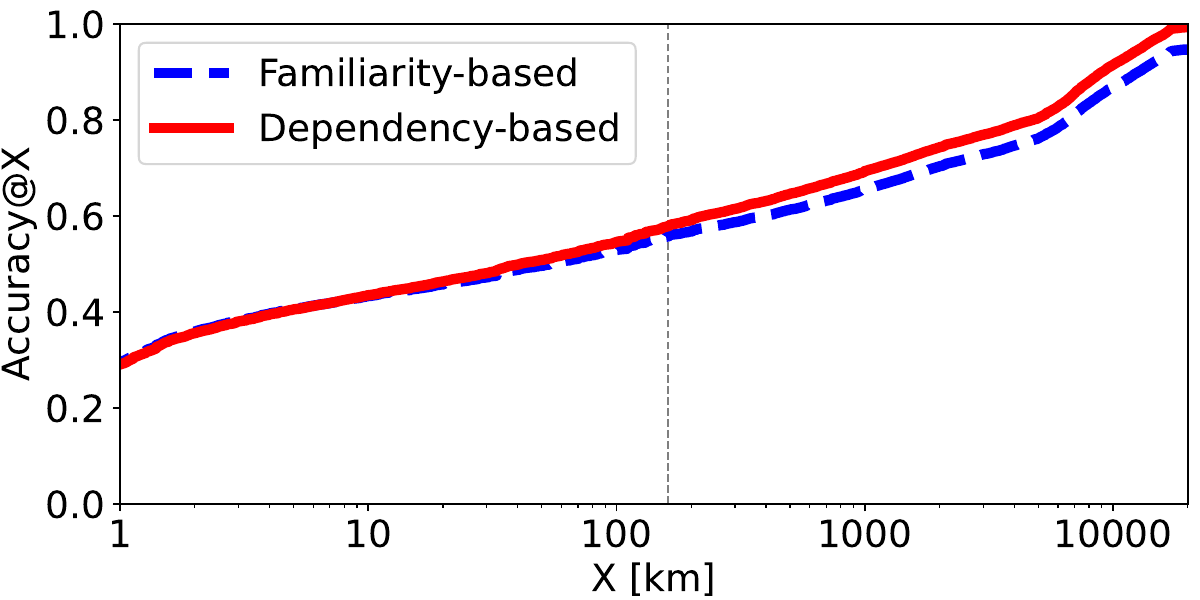}
\caption{Geocoding accuracy to tolerance error distance.}
\label{fig-NEL}
\end{center}
\end{figure}

\figref{fig-NEL} shows the results. The accuracy of the dependency-based strategy (0.58) is better than that of the familiarity-based one (0.56). This suggests that choosing the most famous locations does not always work well and that leveraging linguistic contexts by UD helps improve the accuracy of geocoding.

\section{Conclusion}
We have proposed Wikipedia Hyperlink-based Location Linking, a novel method to create a large-scale corpus for geoparsing leveraging hyperlinks. With this method, we have created a new corpus called the WHLL corpus. Each article in the WHLL corpus contains multiple location expressions annotated with coordinates. Our method and corpus enable researchers in this field to (i) train machine learning-based geoparsing models on a large-scale corpus and (ii) tackle geoparsing with ambiguous location expressions.

\section{Limitation}
In our corpus, not all the location expressions are necessarily annotated with coordinates. If a location expression in a Wikipedia page does not have a hyperlink, it is not annotated. Hyperlinks were often omitted for easy-to-understand expressions, such as country names listed alongside city names.

Also, regarding the location expression of the article title, coordinates annotation using \ref{sub:AA} failed when the characters of the article title and those in the text were slightly different. In addition to variations in the writing of Wikipedia page authors, these differences were sometimes due to restrictions on the characters that can be used in article titles. For example, apostrophes in the article title had UTF-8 hexadecimal character code \textsf{27}, but they sometimes appeared as \textsf{CA BC} in the text.

\section{Acknowledgements}
This work was supported by JSPS KAKENHI Grant Number JP	21H04376.

\section{Bibliographical References}
\bibliographystyle{lrec-coling2024-natbib}
\bibliography{lrec-coling2024}

\end{document}